\documentclass[sigconf,10pt]{acmart}

\usepackage{tikz}
\usepackage{amsmath}
\usepackage{graphicx}
\usepackage{booktabs}
\usepackage{xcolor}
\usepackage{subfig}
\usepackage[linesnumbered,ruled,vlined]{algorithm2e}

\usepackage{amsmath}

\SetCommentSty{mycommfont}
\begin{document}

\acmConference[Submitted to review for ACM MM]{}{2024}{USA}

\settopmatter{printacmref=false} 
\renewcommand\footnotetextcopyrightpermission[1]{} 
\pagestyle{plain} 

\title{DMVC: Multi-Camera Video Compression Network aimed at Improving Deep Learning Accuracy}

\author{
    Huan Cui\textsuperscript{1,2}, Qing Li\textsuperscript{3,*}, Hanling Wang\textsuperscript{1}, Yong Jiang\textsuperscript{1} \\
    \textsuperscript{1}Tsinghua University \\
    \textsuperscript{2}Peking University \\
    \textsuperscript{3}Peng Cheng Laboratory
}

\begin{abstract}
  We introduce a cutting-edge video compression framework tailored for the age of ubiquitous video data, uniquely designed to serve machine learning applications. Unlike traditional compression methods that prioritize human visual perception, our innovative approach focuses on preserving semantic information critical for deep learning accuracy, while efficiently reducing data size. The framework operates on a batch basis, capable of handling multiple video streams simultaneously, thereby enhancing scalability and processing efficiency. It features a dual reconstruction mode: lightweight for real-time applications requiring swift responses, and high-precision for scenarios where accuracy is crucial. Based on a designed deep learning algorithms, it adeptly segregates essential information from redundancy, ensuring machine learning tasks are fed with data of the highest relevance. Our experimental results, derived from diverse datasets including urban surveillance and autonomous vehicle navigation, showcase DMVC's superiority in maintaining or improving machine learning task accuracy, while achieving significant data compression. This breakthrough paves the way for smarter, scalable video analysis systems, promising immense potential across various applications from smart city infrastructure to autonomous systems, establishing a new benchmark for integrating video compression with machine learning.
\end{abstract}



\maketitle

\section{Introduction}


Video analysis systems currently face significant challenges due to the massive volumes of video data, including issues related to transmission, storage, and analysis. The key to addressing these challenges lies in efficient video compression technologies. With video data rapidly dominating internet traffic, the need for compact video representation has never been more critical. To meet this demand, researchers have developed a range of video coding and decoding standards, such as H.264/AVC \cite{1218189}, H.265/HEVC \cite{sullivan2012overview}, and H.266/VVC \cite{bross2021overview}, alongside a series of deep learning-based neural encoders and decoders \cite{gao2020low, sheng2022temporal}. These innovations aim to enhance the rate-distortion (RD) performance of video compression, crucial for both human and machine's consumption of video content. Advances in deep learning-based machine vision have broadened video data's application scope, making it indispensable in machine analysis tasks \cite{guo2021crossroi, zhang2015design}. In applications ranging from online meetings to autonomous driving and smart cities, both humans and machines rely on decoded video information for various purposes—humans for viewing and machines for conducting a plethora of analysis tasks \cite{ling2021rt,yang2021deeprt}. This dual requirement necessitates video content that is not only visually pleasing to humans but also conducive to machine processing, highlighting the differing demands for video information between humans and machines. Humans prioritize visual quality, while machines require precision in deep learning features for enhanced task performance \cite{liu2019caesar, jiang2018chameleon}. This discrepancy introduces a new research direction: designing video compression technologies that cater to both human visual quality and machine analysis needs for extensive, multi-camera video feeds.

\begin{figure}[t]
  \centering
  \includegraphics[width=0.97\linewidth]{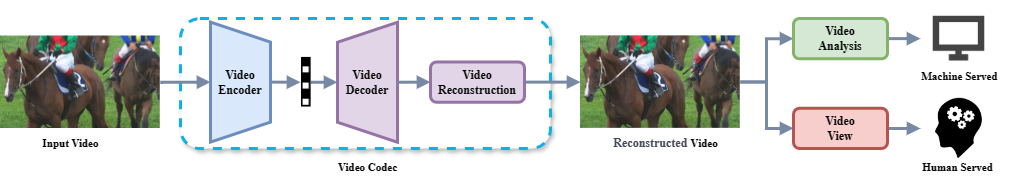}
  \caption{Basic framework of traditional video compression and analysis system}
  \label{fig:chuantongkj}
\end{figure}

While there's a rich body of research on video compression frameworks meeting human visual quality, exploration into systems designed for machine precision is ongoing. The sheer volume of video data necessitates continuous analysis by machines, especially in multi-camera setups. Adhering to traditional video compression and transmission standards could generate an unsustainable amount of data, severely impacting the efficiency of video analysis systems. Furthermore, the divergent needs of machines and human vision complicate the adaptability of restored video content for machine purposes. Traditional compression frameworks, as depicted in Figure \ref{fig:chuantongkj}, can drastically reduce the accuracy of machine learning tasks and consume excessive resources during compression, transmission, and reconstruction.

\begin{figure}[t]
  \centering
  \includegraphics[width=0.97\linewidth]{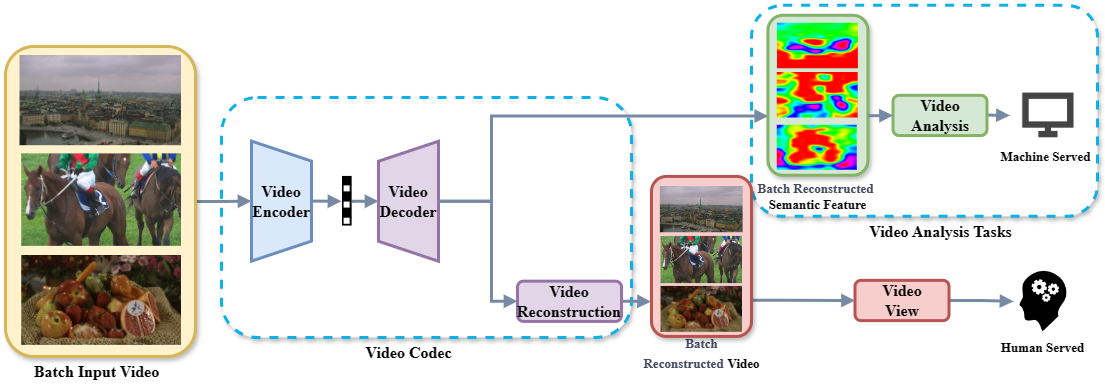}
  \caption{DMVC Basic Framework}
  \label{fig:wodekj}
\end{figure}

To tackle these challenges, we introduce DMVC, an innovative video coding framework tailored for video analysis systems handling multi-stream video feeds. DMVC strategically separates semantic information from human visual features in videos, targeting machine precision while accommodating human viewership and maximizing overall system efficiency. As illustrated in Figure \ref{fig:wodekj}, DMVC performs batch inference on multiple video frames and encodes and decodes multi-stream video bitstreams at the entropy model stage, leveraging potential temporal and spatial correlations between streams to compress data more effectively.

Our contributions are outlined as follows:
\begin{itemize}
\item We introduce a video compression network optimized for machine precision in deep learning tasks, emphasizing the preservation of crucial semantic information relevant to these tasks. This approach not only maintains video quality but also significantly improves the precision of deep learning tasks.
\item We propose a mechanism for compressing semantic information that focuses on retaining essential data for deep learning tasks, enabling more efficient use of storage space and bandwidth. This mechanism also facilitates a lightweight frame reconstruction process, thereby reducing the complexity of coding and decoding.
\item We present a scalable and adaptable video compression network designed to be flexible across various deep learning tasks and video data types in multi-stream scenarios. The network's architecture and methodologies can be adjusted and extended to suit the diverse requirements of different tasks and environments.
\end{itemize}

\section{Related Works}
\subsection{Human-centric Video Compression}

Decades of development in traditional video compression technology have led to the development of various video coding standards. Among them, the H.264/AVC standard, developed by ITU-T and ISO/IEC between 1999 and 2003, is widely used in high-definition television broadcasting, internet videos, and mobile network videos. The introduction of the H.265/HEVC standard \cite{sullivan2012overview} in 2013, which offers a bitrate reduction of about 50\% compared to H.264/AVC \cite{1218189}, leveraged advancements in video resolution and parallel processing technologies. Further, the H.266/VVC standard, the latest in video coding, significantly lowers bitrates compared to H.265/HEVC to meet the demands of both current and emerging media. These standards share a hybrid video coding framework that includes stages like prediction, transformation, quantization, entropy coding, and loop filtering\cite{taki1974interframe,6768003}.

The rise of neural network-based codecs \cite{jain2018rexcam,4425348,yan2018convolutional, alam2015perceptual}, primarily relying on residual coding, marks a recent innovation. A groundbreaking study \cite{lu2019dvc} replaced traditional codec components, such as motion estimation and compensation, with neural networks, optimizing them on an end-to-end basis. Hu et al. \cite{5652119} advanced pixel-level prediction and reconstruction to feature level. Rippel et al. \cite{10153603} introduced a flexible rate control specifically for deep video coding. Beyond residual coding, the shift towards conditional coding leverages temporal features as conditions for compressing current frames, with further enhancements in rate-distortion (RD) performance achieved through feature propagation and multi-scale spatio-temporal backgrounds \cite{lu2020end}. Notably, scalable coding through the neural network-based Swift \cite{vaswani2017attention} scheme enables scalable video coding optimized for human vision, without the need for cross-layer references.

These algorithms, while focusing on human visual quality, might compromise video analysis performance due to decreased reconstruction quality. Thus, our research prioritizes efficient compression of machine-analyzable video features, significantly improving rate-accuracy performance. Inspired by scalable coding, our work achieves seamless adaptability from machine to human vision.

\subsection{Machine-centric Video Compression}

Deep learning-based video compression has emerged as a vibrant research area \cite{chen2019learning, srivastava2015unsupervised, cui2018convolutional, li2017convolutional,8554306, pfaff2018neural, 4703262}. Lu et al. \cite{lu2020end} enhanced compression efficiency by replacing traditional video compression components with CNNs, optimizing the rate-distortion cost across the entire network. Lin et al. \cite{lin2020m} minimized motion vector coding costs by generating more accurate current frame predictions using multiple reference frames and their motion vectors. Yang et al. \cite{yang2020learning} introduced a novel recursive learning video compression approach that utilizes cross-frame temporal information for latent representation and compressed output reconstruction. Habibian et al. \cite{habibian2019video} proposed a 3D AutoEncoder for direct video compression, while Liu et al. \cite{119325} explored frame-to-frame temporal correlations using separate image codecs for each frame and entropy models.

These deep learning approaches have paved new paths for machine feature-based video compression. To boost the efficiency of machine vision task, recent learning-based methods \cite{matsubara2022supervised,minnen2020channel} aim for joint optimization of feature compression and task analysis. However, restoring high-quality videos from compact features that also meet human viewing requirements poses a significant challenge. Various coding schemes have been proposed to cater to both machine and human visual needs. Huang \cite{huang2022hmfvc} proposed extracting semantic information from motion flows for both machine analysis and signal reconstruction. Some studies \cite{yi2022task} have fine-tuned task networks for analyzing and reconstructing the same bitstream, ensuring videos are suitable for human viewing and machine analysis. Different from single-bitstream methods, other research \cite{10123027, 10228933} employs additional bitstreams for analysis, proposing scalable coding schemes that use base layer features for machine analysis and enhancement information for human visual reconstruction.

While these efforts offer valuable insights into developing video compression and analysis systems that consider both human visual quality and machine analysis precision, most have focused on transmitting additional semantic information alongside video reconstruction, without fully separating features for human and machine analysis. Moreover, they have not fully addressed the specific needs of multi-stream video compression scenarios. Our work concentrates on designing and utilizing machine semantic features to minimize redundant information transmission and enhance compression efficiency in multi-stream video scenarios.

\section{Overall Design}
In this section, we detail the DMVC design, starting with an overview of its architecture. Subsequently, we delve into the functionalities and implementation specifics of each system module. Finally, the model's training intricacies are presented.

\subsection{Architecture}

\begin{figure}[t]
\centering
\includegraphics[width=0.97\linewidth]{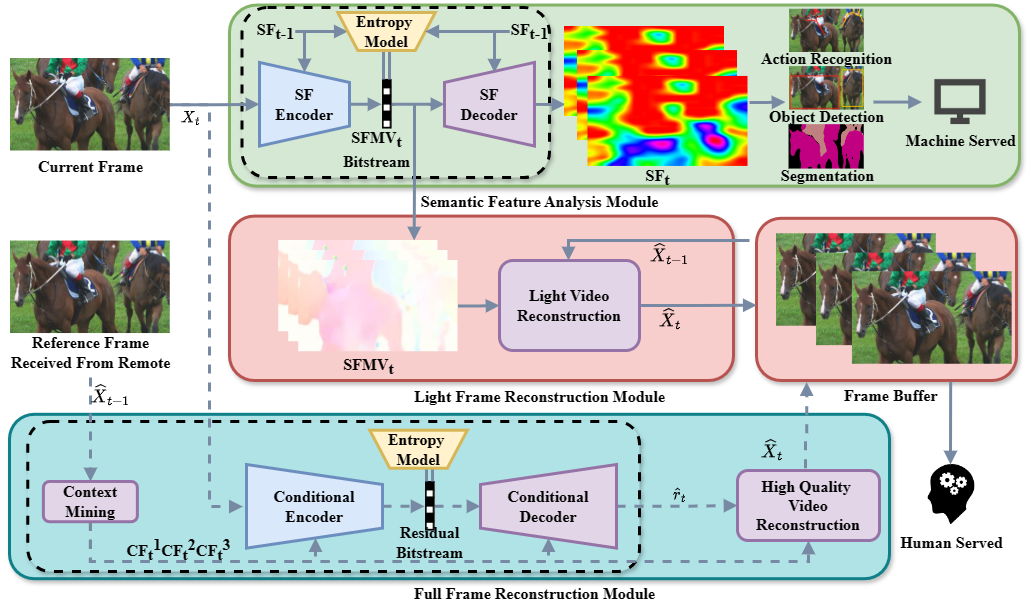}
\caption{DMVC's Comprehensive Architecture}
\label{fig:dmvc_architecture}
\end{figure}

DMVC, as depicted in Figure \ref{fig:dmvc_architecture}, comprises three primary modules: the Semantic Feature Analysis Module, the Lightweight Video Frame Reconstruction Module, and the Full Frame Reconstruction Module. Their respective roles are outlined as follows:

\textbf{Semantic Feature Analysis Module} is dedicated to machine video analysis tasks. It processes video frames to encode and decode advanced machine semantic information, facilitating machine analysis tasks. By employing a conditional context encoder-decoder, the module compresses semantic features to reduce their encoding bitrate. Notably, it seeks to capture and reconstruct transformations between current and reference semantic features, analogous to the optical flow in motion estimation. Initially, it combines the current frame $X_t$ with the prior context feature $SF_{t-1}$, calculating semantic transformation information $SFMV_t$ across multiple scales. This semantic transformation data, alongside $SF_{t-1}$, are then fed into a decoding network to reconstruct $SF_t$. Subsequently, $SF_t$ is used for machine analysis. The vision analysis task module adapts by integrating $SF_t$ for task-specific training. Additionally, $SFMV_t$ directly facilitates the lightweight reconstruction of video frames.

\textbf{Lightweight Video Frame Reconstruction Module} caters to human viewing needs by fulfilling the requirements for standard quality frames. It leverages $SFMV_t$ to predictively reconstruct frames. Specifically, it transforms and predicts the current frame $\hat{X}{t}$ from $SFMV_t$ and a reference frame $\hat{X}{t-1}$, achieving efficient frame reconstruction. This module's dependency on remote inputs and structures obviates the need for local storage of reconstructed frames, thereby decoupling semantic from video reconstruction information and significantly conserving edge resources. This also enables parallel execution of frame reconstruction and video analysis tasks remotely.

\textbf{Complete Video Frame Reconstruction Module} is activated upon demand for high-quality frames, such as in detailed scene examinations. Unlike the lightweight module, which satisfies general viewing requirements, this module is not always active, hence its depiction with a dashed line. It enhances the quality of reconstructed frames starting from lightweight ones, $\hat{X}{t}$. This involves requesting $\hat{X}{t}$'s copy from the remote to the edge, extracting multi-scale contextual features for high-quality reconstruction.

\subsection{Detailed Module Introduction}

This segment provides an in-depth overview of the functionality and implementation specifics of each module within the system.

\subsubsection{Semantic Feature Analysis Module}

This module performs semantic-level compression of semantic features to aid machine analysis. Given the substantial similarity between consecutive video frames—more pronounced within the high-level semantic feature maps—this similarity can be harnessed to reduce the encoding bitrate for semantic features. There are several methodologies for compressing semantic features, such as internal coding techniques, traditional predictive coding paradigms, or conditional coding methods. Applying internal coding directly to semantic features disregards temporal correlations. Traditional predictive coding approaches require additional bitstreams, like optical flow. Conversely, conditional coding utilizes temporal context as a condition to autonomously explore spatial-temporal correlations. Compared to residual coding, conditional coding offers lower or equivalent entropy limits. Our approach encodes semantic features directly, obviating the need for supplementary predictions. Moreover, these features can exploit spatial-temporal correlations to further decrease encoding bits. Additionally, due to the variance in object sizes within the video field of vision, multi-scale semantic features exhibit superior representational efficacy. Thus, we've designed the SF Encoder-Decoder network, inspired by conditional coding principles.

The module's architecture is detailed in Figure \ref{fig:mokuai1}. Leveraging \(SF_{t-1}\)'s rich, high-dimensional channel information as a condition, we aim to minimize spatial-temporal redundancy in semantic features. In practice, \(X_t\) and \(SF_{t-1}\) are concatenated and fed into the SF Encoder, generating multi-scale semantic transformation \(SFMV_t\), which is then compacted into a bitstream by an entropy model. Subsequently, the SF Decoder reconstructs the initial semantic feature \(SF_{t}\) with assistance from \(SF_{t-1}\), as depicted in equation \ref{eq:mokuai11}:

\begin{figure}[h]
  \centering
  \includegraphics[width=1.0\linewidth]{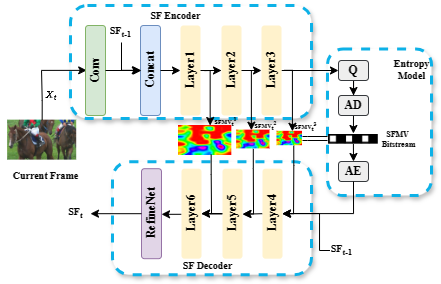}
  \caption{Semantic Feature Analysis Module Structure}
  \label{fig:mokuai1}
\end{figure}

\begin{equation}
\begin{aligned}
{SF}_{t} = Dec(Enc(X_t|SF_{t-1})|SF_{t-1})
\end{aligned}
\label{eq:mokuai11}
\end{equation}

\noindent Here, \(Enc\) and \(Dec\) denote the SF Encoder-Decoder's semantic encoder and decoder, respectively, excluding any refinement module. For the inaugural P-frame, its decoded reference frame (I-frame) is inputted into the task analysis network to acquire \(SF_{t-1}\).

Post-decoding of semantic features, they are input into a refinement module to counteract quantization errors, according to equation \ref{eq:mokuai12}:

\begin{equation}
\begin{aligned}
 & \hat{SF}_{t} = {SF}_{t} + \alpha_t \cdot {SF}_{t}, \\
 & \alpha_t = Softmax\left(\frac{1}{N} \sum_{i=0}^{N-1} Layer({SF}_{t}) \cdot Layer(SF_{t-i-1})\right),
\end{aligned}
\label{eq:mokuai12}
\end{equation}

\noindent Wherein \(N\) denotes the count of previously decoded semantic features. \(\alpha_t\) represents the aggregation weight. \(Layer(\cdot)\) signifies a feature extraction module incorporating two convolutional layers.

This approach ensures the Semantic Feature Analysis Module not only efficiently compresses but also reconstructs semantic information, utilizing antecedent knowledge and minimizing redundancy through sophisticated conditional encoding and refinement techniques.

\subsubsection{Video Reconstruction Module}

\begin{figure}[h]
  \centering
  \includegraphics[width=1.0\linewidth]{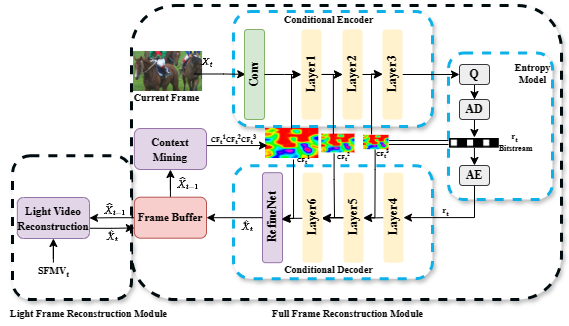}
  \caption{Video Reconstruction Module Structure}
  \label{fig:mokuai2}
\end{figure}

The intricacies of the Video Reconstruction Module are depicted in Figure \ref{fig:mokuai2}.

Drawing inspiration from traditional scalable video encoding schemes, where videos are encoded into multiple layers each representing a different quality aspect of the video scene. The base layer represents the lowest quality level, and one or more enhancement layers are encoded by referencing the lower layers. In these traditional schemes, inter-layer references and predictive tools leverage information from lower layers to improve the rate-distortion (RD) efficiency of the enhancement layers.

Motivated by this, we initially developed a Lightweight Video Reconstruction Module, which utilizes the semantic transformation information \(SFMV_t\) and receives reference frames \(\hat{X}_{t-1}\) from a historical frame cache. By leveraging \(SFMV_t\), it predicts and transforms \(\hat{X}_{t-1}\) to generate the current frame's predicted frame \(\hat{X}_{t}\), achieving a lightweight reconstruction of the current frame predictively, as shown in equation \ref{eq:mokuai21}.

\begin{equation}
\begin{aligned}
 & \hat{X}_t = Warp(Conv(\hat{X}_{t-1}), \hat{SFMV}_{t}), \\
\end{aligned}
\label{eq:mokuai21}
\end{equation}

\noindent Here, \(Conv(\cdot)\) represents a convolution operation.

Furthermore, we've designed a Full Video Frame Reconstruction Module to cater to scenarios where there's a demand for high-quality frames, such as detailed inspection or evidence review. This module makes use of a copy of the reference frame \(\hat{X}_{t}\) and employs a context miner to extract multi-scale contextual features, denoted as \({CF_t}^1{CF_t}^2{CF_t}^3\). These multi-scale contextual features serve as conditions for encoding the residual \(\hat{r}_{t}\) between the lightweight reconstructed frame and the current frame, as illustrated in equation \ref{eq:mokuai22}:

\begin{equation}
\begin{aligned}
 & \hat{r}_t = Warp(Conv({X}_{t}), {CF_t}^1{CF_t}^2{CF_t}^3), \\
 & \hat{X}_t = Refine(\hat{r}_{t}, {CF_t}^1{CF_t}^2{CF_t}^3),
\end{aligned}
\label{eq:mokuai22}
\end{equation}

\noindent Where \(Refine(\cdot)\) denotes the Refinenet’s refinement of the reconstructed frame’s quality.

This architectural approach ensures that the Video Reconstruction Module not only efficiently predicts and reconstructs frames based on semantic transformations but also adapts to varying quality demands through a scalable encoding strategy, thus providing a versatile solution for both routine viewing and high-quality frame analysis requirements.

\subsection{Training Details}
DMVC is composed of multiple modules, and we have implemented a hierarchical training approach for the video compression network.
It's important to note that, unlike previous semantic feature compression networks, our semantic feature compression module is designed to extract and compress high-level semantic feature transformations \(SFMV_t\), akin to optical flow. Therefore, we first train the SF Encoder-Decoder structure within the Lightweight Frame Reconstruction Module along with Light Video Reconstruction. The aim is to ensure that under the influence of \(SFMV_t\), the predicted frames produced by the lightweight frame reconstruction module are of similar quality to those predicted using optical flow methods and the original frames. The loss function used during this training process is as follows in equation \ref{eq:xijie1}:

\begin{equation}
\begin{aligned}
 & L_sf = R_sf + \lambda_1 D_sf, \\
 & D_sf = D(\hat{X}_t, X_t),
\end{aligned}
\label{eq:xijie1}
\end{equation}

\noindent Where \(R_sf\) is calculated based on the encoding bitrate of \(SFMV_t\), \(\mathrm{D}(\cdot)\) denotes frame-level distortion, calculated using MSE and MS-SSIM. \(\lambda_1\) balances the trade-off between compression bitrate and the quality of the frame reconstructed using semantic transformation information.

Subsequently, with the SF Encoder-Decoder structure and Light Video Reconstruction in the lightweight frame reconstruction module fixed, we use the reconstructed semantic features obtained from the SF Decoder to train the video analysis task network, as depicted in equation \ref{eq:xijie2}:

\begin{equation}
\begin{aligned}
 & L_v = MSE(F_T(X_t), F'_T(SF_{t})) + \beta_1 L_{task},
\end{aligned}
\label{eq:xijie2}
\end{equation}

\noindent Here, MSE stands for Mean Squared Error, \(F_T(\cdot)\) represents the original backbone network of the analysis task. \(F'_T(\cdot)\) signifies the modified version, that is, the version of the video analysis task network combined with DMVC after removing the feature extraction module. \(\beta_1\) moderates the compromise between compression bitrate and task analysis precision. \(L_{task}\) indicates the machine analysis loss.

Following that, with the above weights fixed, we train the multi-scale context conditional encoder and decoder and the context feature miner in the complete frame reconstruction module, as shown in equation \ref{eq:xijie3}:

\begin{equation}
\begin{aligned}
 & L_c = R_c + \lambda_2 D_c, \\
& D_c = D(\hat{X}_t, X_t),
\end{aligned}
\label{eq:xijie3}
\end{equation}

\noindent Where \(R_c\) is determined by the encoding bitrate of \(\hat{r}_t\), \(\mathrm{D}(\cdot)\) represents frame-level distortion, evaluated using MSE and MS-SSIM. \(\lambda_1\) adjusts the balance between compression bitrate and the quality of the frame reconstructed by the multi-scale context conditional decoder.

This training regimen enables DMVC to efficiently handle both semantic feature compression for machine analysis and high-quality frame reconstruction for human viewing, leveraging the sophisticated relationships between semantic transformations, compression efficiency, and reconstruction fidelity across its modules.

\section{Experimental Evaluation}

\subsection{Experimental Setup}

\textbf{Dataset} To assess our method, we focused on video object detection as the visual task, utilizing seven parts of the Nuscenes dataset. We implemented the DMVC visual compression and analysis system on the Nuscenes dataset, adhering to the training and testing configurations defined in MMtracking and MMaction2. Additionally, we evaluated the video reconstruction performance on the HEVC Common Test Conditions (CTC) to showcase DMVC's capability in video frame reconstruction.

\textbf{Evaluation Metrics} We employed Mean Average Precision (mAP) to assess the performance of the machine vision task and PSNR and MS-SSIM to measure video reconstruction quality. The encoding cost was evaluated using bits per pixel (bpp).

\textbf{Comparison Setup} Our method was compared against traditional codecs like x264, x265, and popular neural network-based codecs like DVC.

This comprehensive experimental design and evaluation framework takes into account both the performance of machine vision tasks, such as the accuracy of video object detection, and human visual requirements, like the quality of video reconstruction. Comparisons with traditional and cutting-edge codecs demonstrate the advantages of our approach.

\subsection{Experimental Results}
\subsubsection{Runtime Cost Performance}

\begin{figure}[h]
  \centering
  \includegraphics[width=0.5\linewidth]{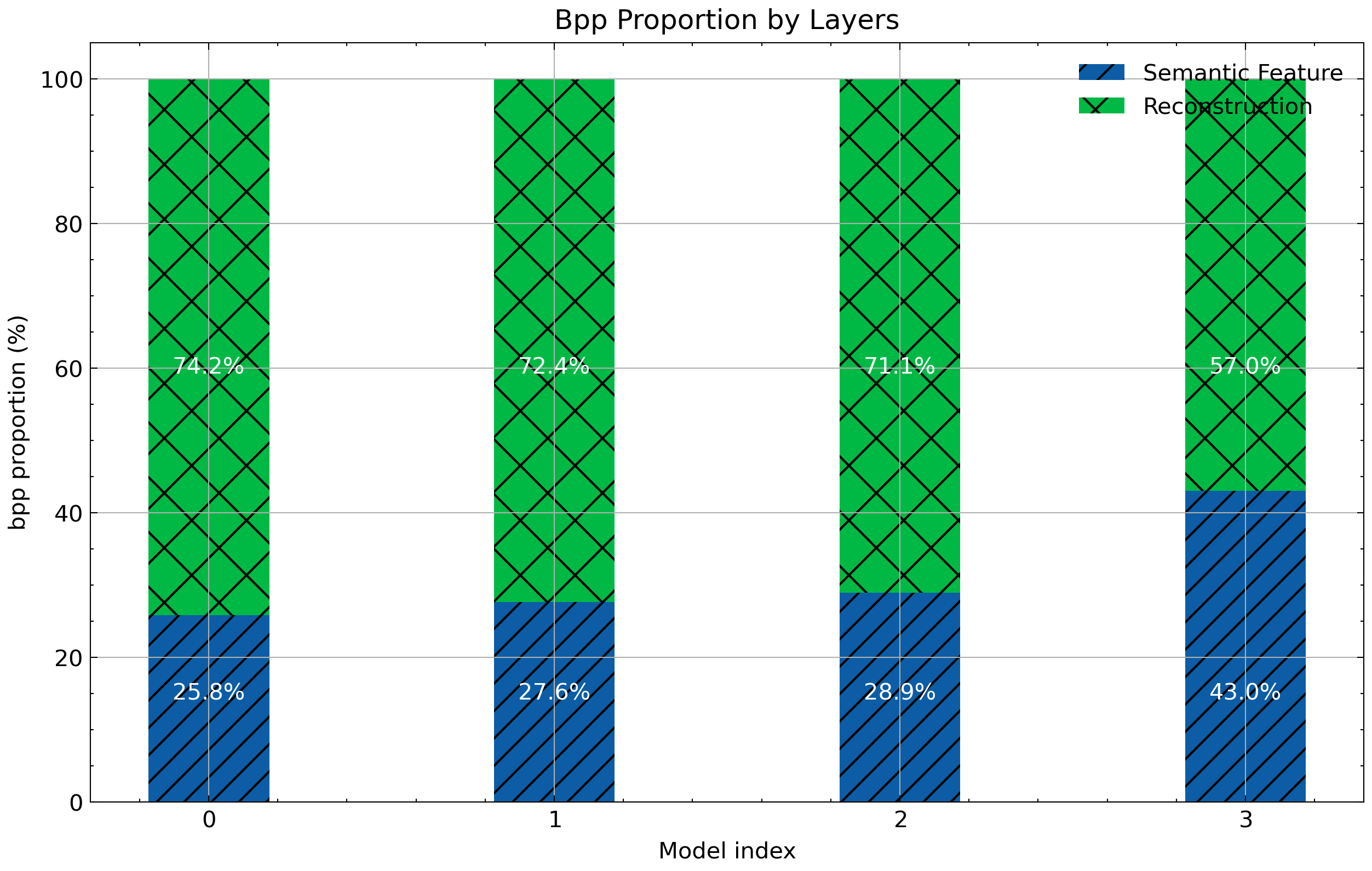}
  \caption{Bitstream}
  \label{fig:code}
\end{figure}

We first analyze the bitrate of DMVC applied on the HEVC dataset. As shown in Figure \ref{fig:code}, it is evident that within various encoding layers, the semantic feature compression layer consumes less data, making it the most lightweight in terms of residual data compressed by the reconstruction layer. Moreover, as compression efficiency improves, the data volume of the semantic feature layer significantly increases. In contrast, the semantic feature layer requires relatively less data, indicating its efficiency during the encoding process. Specifically, in scenarios where video reconstruction is not pursued, transmitting only the essential bitstream for analysis can substantially reduce the required bandwidth.

Table \ref{tab:time} compares the encoding and decoding time performance of the benchmark neural codec DCVC and our method DMVC. It is evident that DMVC far outperforms DCVC in terms of encoding and decoding time, especially in decoding time where DMVC is over 200 times faster than DCVC. This is crucial for applications that require rapid processing of large volumes of video data. Moreover, the significant reduction in encoding time also makes DMVC more practical for real-time video processing and streaming services. These advantages demonstrate that DMVC is a powerful and efficient video codec solution.

\begin{table}
  \centering
  \caption{Time}
  \begin{tabular}{clclclclc}
    \toprule
    Video     & \multicolumn{2}{c}{Encoding Time(s)}&   \multicolumn{2}{c}{Decoding Time(s)}  \\
    Codec&Module1 &Module2 &Module1 &Module2 \\
    \midrule
    DCVC & \multicolumn{2}{c}{3.80}&   \multicolumn{2}{c}{21.07} \\
    *DMVC & 0.10 &  0.39 & 0.10 & 0.9 \\
    \bottomrule
  \end{tabular}
  \label{tab:time}
\end{table}

\subsubsection{Accuracy Performance}

\begin{figure}[h]
  \centering
  \includegraphics[width=1.0\linewidth]{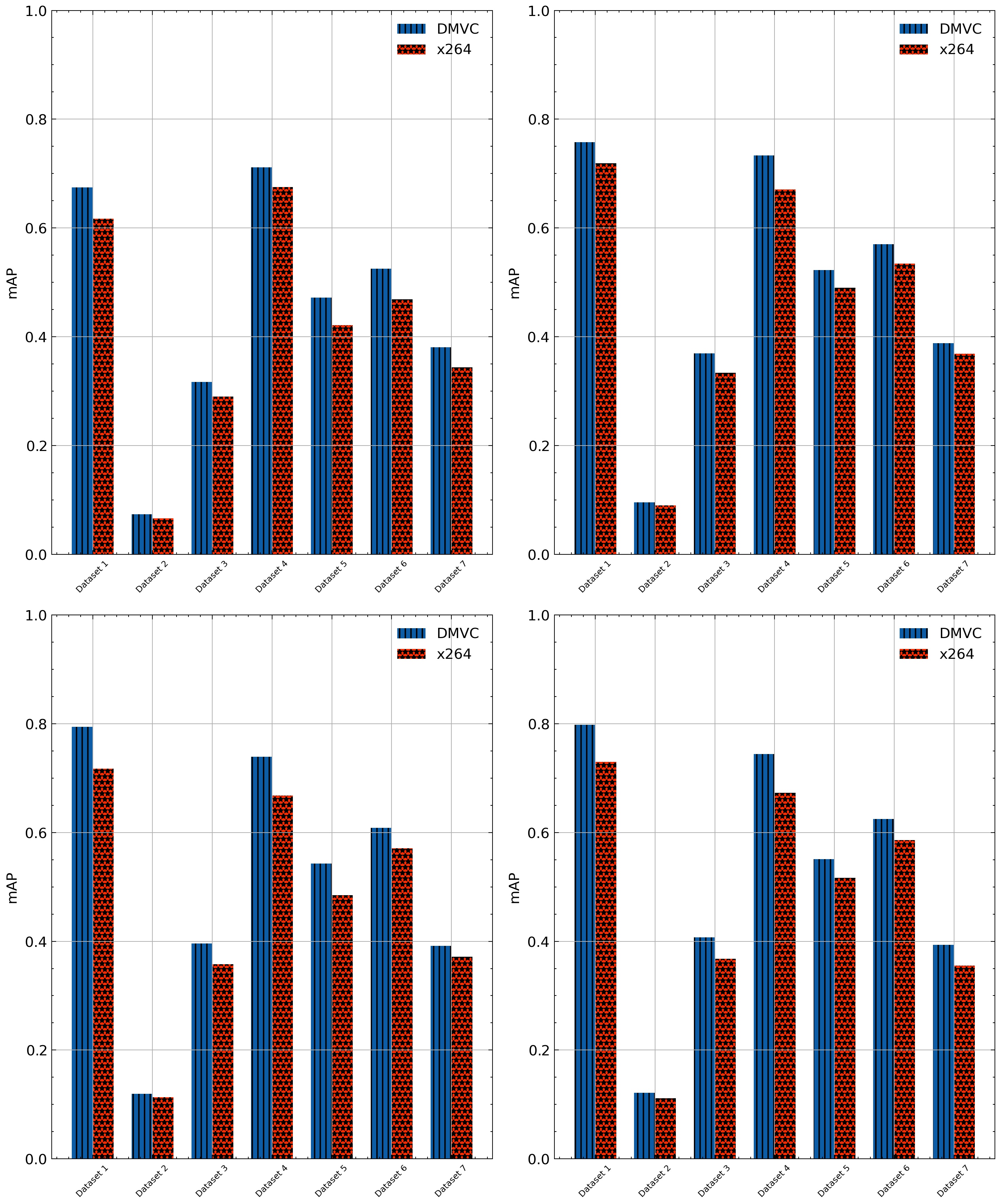}
  \caption{Nuscenes Results}
  \label{fig:nu}
\end{figure}

In Figure \ref{fig:nu} and Table \ref{tab:top1}, we detailed the performance comparison between DMVC and traditional codecs like x264 in executing object detection tasks on the Nuscenes dataset. Notably, DMVC achieved superior detection performance with fewer bits transmitted, highlighting the effectiveness of our techniques for compressing and extracting higher-order semantic features and the excellent performance of DMVC modules in video coding-decoding and feature extraction. This further proves DMVC's advanced and practical value in the field of video compression. Through careful design and optimization of each module, DMVC achieves higher data efficiency and better performance in multi-stream video compression processes. Compared to traditional codecs like x264, DMVC not only significantly reduces the required transmission bandwidth but also maintains or improves accuracy in advanced video analysis tasks like object detection. This indicates that leveraging DMVC's efficiency in processing video data allows for high-quality video surveillance and analysis under bandwidth-limited conditions, providing strong technical support for applications such as autonomous driving and city surveillance. Furthermore, DMVC's breakthrough also introduces a new research direction in video coding-decoding technology: maximizing the practical value and analysis performance of video content while minimizing data transmission.

\begin{table}
  \centering
  \caption{Nuscenes 2 Results}
  \begin{tabular}{lclclclcl}
    \toprule
    Dataset     & x264&   x265&  DCVC   &  *DMVC  \\
    \midrule
    Nuscenes &  72.5\% & 76.0\% & {\color{red} 58.6\%} & {\color{blue} 79.4\%} \\
    \bottomrule
  \end{tabular}
  \label{tab:top1}
\end{table}

\subsubsection{Rate-Distortion Performance}

\begin{figure}[h]
	\centering  
	\subfloat{
		\includegraphics[width=0.9\linewidth]{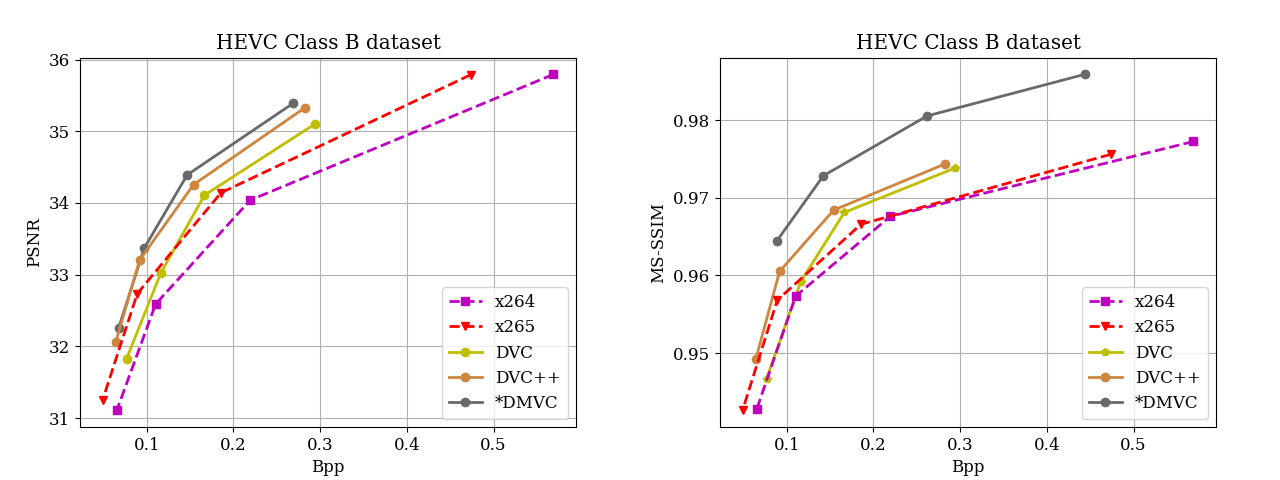}}
  \hspace{0.1em}
	\subfloat{
		\includegraphics[width=0.9\linewidth]{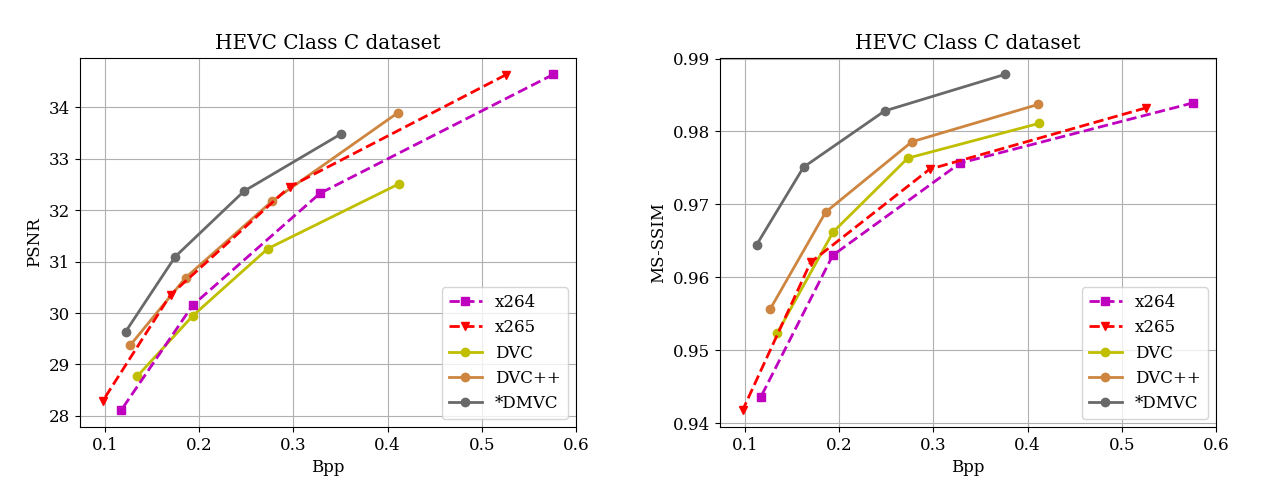}}
  \hspace{0.1em}
  \subfloat{
		\includegraphics[width=0.9\linewidth]{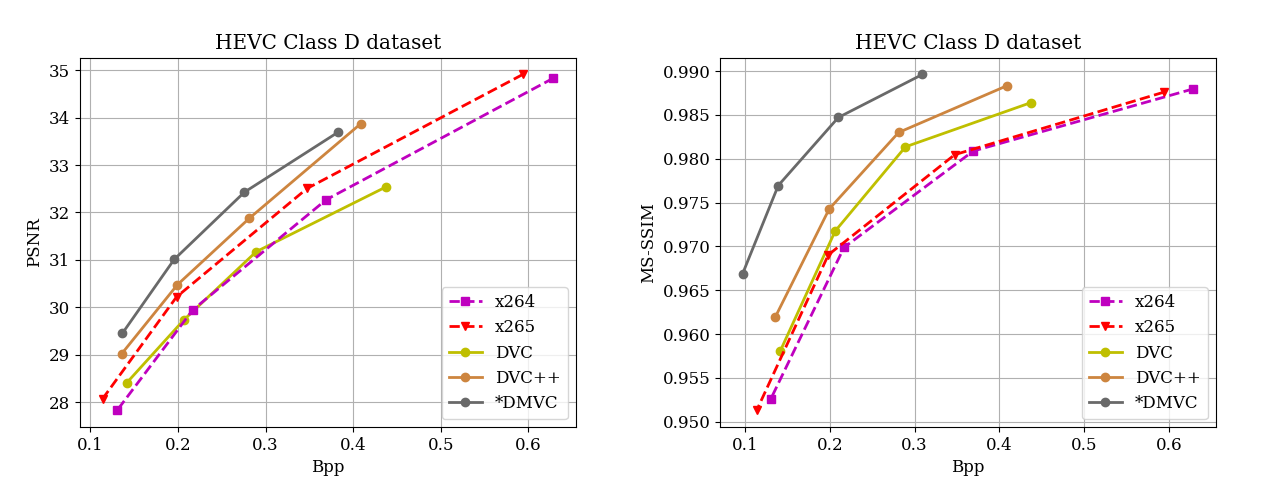}}
  \hspace{0.1em}
  \subfloat{
		\includegraphics[width=0.9\linewidth]{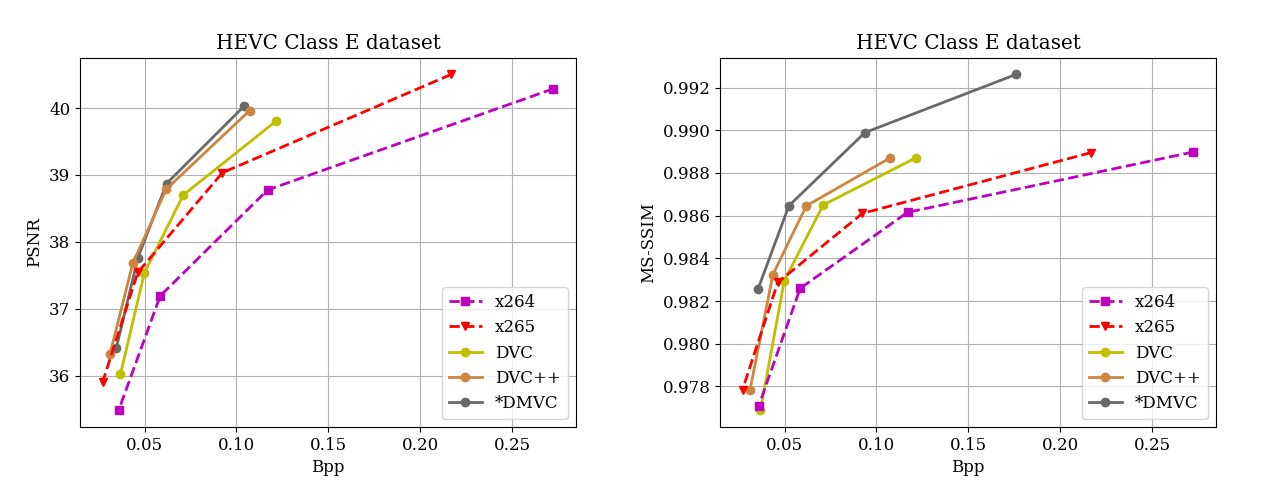}}
        \caption{HEVC Results}
        \label{fig:hevc}   
\end{figure}

In Figure \ref{fig:hevc}, we showcase the rate-distortion performance visually, illustrating the relationship between PSNR, MS-SSIM, and bitrate (bpp). The encoding cost calculation is based on the total bitstream generated within different modules of our DMVC framework, including the temporal-spatial motion vector \(SFMV_t\) and the reconstruction residual \(\hat{r}_t\). Remarkably, our proposed method exhibits highly competitive performance at lower bitrates, as evidenced by significant improvements in both key metrics, PSNR, and MS-SSIM. This achievement reveals an important insight: even under more compact bitstream conditions, our method can maintain the visual quality of video content with minimal impact on the viewer's visual experience by effectively compressing and precisely extracting higher-order semantic features. This advantage is significant when compared to current mainstream codecs, such as x264, x265, and even recently popular deep-learning-based solutions like DVC. Whether evaluated by PSNR or MS-SSIM, DMVC demonstrates a clear advantage in maintaining visual quality and compression efficiency, especially in high-compression scenarios, effectively balancing rate-distortion performance.

\section{Conclusion}

This paper primarily introduces the DMVC model, designed to enhance the performance of deep learning tasks by integrating them with precision. Unlike traditional video compression technologies primarily optimized for human visual quality, a significant advantage of DMVC lies in its ability to ensure high accuracy in deep learning tasks while compressing video. This enhancement in accuracy brings significant benefits across various application scenarios, including video analysis and behavior recognition. Moreover, by focusing on preserving information crucial for deep learning tasks, the model operates more efficiently in terms of storage space and bandwidth usage. This aspect is particularly valuable in environments where storage costs are high or network resources are limited. DMVC also places a strong emphasis on scalability and adaptability, indicating that its design is sufficiently flexible to adjust according to different deep learning tasks and types of video data. Its architecture and techniques, suitable for large-scale, multi-channel video analysis, can be optimized and adjusted according to varying task requirements and application scenarios, achieving a balance between serving machine tasks and maintaining human visual perception quality.

\bibliographystyle{ACM-Reference-Format}
\bibliography{refs}

\end{document}